\documentclass{article}

\PassOptionsToPackage{numbers, compress}{natbib}

\usepackage[final]{new_in_ML}
\usepackage{graphicx}
\usepackage{amsmath}
\usepackage[utf8]{inputenc} 
\usepackage[T1]{fontenc}    
\usepackage{hyperref}       
\usepackage{url}            
\usepackage{booktabs}       
\usepackage{amsfonts}       
\usepackage{nicefrac}       
\usepackage{microtype}      
\usepackage{xcolor}         
\usepackage{natbib}
\usepackage{floatrow}
\newfloatcommand{capbtabbox}{table}[][2\FBwidth]
\usepackage{blindtext}
\usepackage{multirow}

\title{Optimizing Inventory Routing: A Decision-Focused Learning Approach using Neural Networks}

\author{MD Shafikul Islam, Azmine Toushik Wasi
\\
Shahjalal University of  Science and Technology\\
\texttt{\{Email:shafikul37@student.sust.edu,azminetoushik.wasi@gmail.com\}}}

\begin{document}

\maketitle

\begin{abstract}
Inventory Routing Problem (IRP) is a crucial challenge in supply chain management as it involves optimizing efficient route selection while considering the uncertainty of inventory demand planning.
To solve IRPs, usually, a two-stage approach is employed, where demand is predicted using machine learning techniques first, and then an optimization algorithm is used to minimize routing costs.Our experiment shows machine learning models fall short of achieving perfect accuracy because inventory levels are influenced by the dynamic business environment, which, in turn, affects the optimization problem in the next stage, resulting in sub-optimal decisions. 
In this paper, we formulate and propose a decision-focused learning-based approach to solving real-world IRPs.This approach directly integrates inventory prediction and routing optimization within an end-to-end system potentially ensuring a robust supply chain strategy.


\end{abstract}

\vspace{-3mm}
\section{Introduction}
\vspace{-3mm}
The inventory routing problem integrates inventory management and vehicle routing decisions, improving supply chain solutions by jointly addressing inventory levels and customer delivery routing \cite{campbell1998inventory}.
Various approaches, including branch and cut, genetic algorithms, and heuristics/meta-heuristics, aim to find optimal or near-optimal solutions in most models. In the dynamic business environment, where demand fluctuations significantly affect overall profit, few models focus on integrating demand prediction with optimal decision-making. To address uncertainty, the problem is typically tackled using a two-stage method. In the first stage, demand prediction is performed, and in the second stage, the predicted value is used in an optimization algorithm to make decisions \cite{vanderschueren2022predict}.
But enhancing prediction accuracy doesn't always guarantee optimal decision-making \cite{wilder2019melding}. Moreover, in the dynamic business landscape, improving accuracy can be challenging, often resulting in sub-optimal decisions.
Decision-focused learning (DFL) takes a unique approach by not segregating predictive modeling and optimization. Instead, it integrates them by connecting the machine learning model directly to decision quality. In DFL, the loss function relies on the solution of an optimization model, effectively embedding the optimization solver as part of the machine learning model, leading to more effective decision-making \cite{wilder2019melding}.


\vspace{-5mm}
\section{Inventory Routing Problem Formulation}
\vspace{-3mm}
For formulating the problem, we introduce several key variables: $I$: Set of customers, $I = \{1, 2, \ldots, N\}$; $T$: Set of time periods, $T = \{1, 2, \ldots, T\}$; $K$: Set of routes, $K$, which may be from the supplier to a customer or between customers; $d_{it}$: Demand of customer $i$ at time $t$; $M_i$: Maximum capacity of customer $i$ to hold inventory; $Q$: Maximum capacity of the vehicle; $h_s$: Holding cost per unit at the supplier; $h_i$: Holding cost per unit at customer $i$, where $i \in I$; $c_{kj}$: Cost of traveling from location $k$ to location $j$, where $k, j \in K$; $x_{ijk}$: Binary variable indicating if route $k$ is used from location $i$ to location $j$; $y_{it}$: Binary variable indicating if customer $i$ is visited at time $t$; $s_{it}$: Inventory level of customer $i$ at time $t$; $S_t$: Inventory level at the supplier at time $t$; $P$: Daily production at the supplier; $I_{i0}$: Initial inventory at customer $i$. Decision variable : $q_{it}$: Quantity delivered to customer $i$ at time $t$; $z_{kt}$: Binary variable indicating if route $k$ is used at time $t$.\\
\textbf{Objective Function:} Minimize the total cost of holding inventory and routing:
\begin{equation}
\vspace{-1mm}
\text{Minimize} \quad  \sum_{i \in I} \sum_{t \in T} (h_s \cdot S_t + h_i \cdot s_{it}) + \sum_{t \in T} \sum_{k \in K} c_{kj} \cdot z_{kt}
\end{equation}

\vspace{-7mm}
\subsection{Constraints:}
\vspace{-3mm}
\textbf{Inventory Related Constraints:} 1. Ensure customer demand is met: $\sum_{t \in T} q_{it} = d_{i1}, \quad \forall i \in I$ 2. Inventory level of the customer can't go below zero during visits: $s_{it} \geq 0, \quad \forall i \in I, \forall t \in T$ 3. After receiving a delivery, the customer's inventory can't exceed its capacity: $s_{it} \leq M_i, \quad \forall i \in I, \forall t \in T$.
\textbf{Vehicle Related Constraints:} 1. Total inventory carried in the vehicle cannot exceed its capacity: $\sum_{i \in I} q_{it} \leq Q, \quad \forall t \in T$ 2. Each customer can be visited at most once daily: $\sum_{i \in I} y_{it} \leq 1, \quad \forall t \in T$.
\textbf{Routing Constraints:}  Ensure that only visited customers are on the routes: $\sum_{i \in I} x_{ijk} \geq y_{it}, \quad \forall k \in K, \forall t \in T$.
\textbf{Inventory Balance Constraints:} 1. Inventory balance at the supplier: $S_{t+1} = S_t + P - \sum_{k \in K} \sum_{j \in I} x_{ijk}, \quad \forall t \in T$.
\textbf{Initial Inventory at Customers:} Set the initial inventory at each customer: $s_{i0} = I_{i0}, \quad \forall i \in I$.
\textbf{Vehicle Inventory Constraint:} The total inventory carried by the vehicle must be delivered: $\sum_{i \in I} q_{it} = \sum_{k \in K} \sum_{j \in I} x_{ijk}, \quad \forall t \in T$.

%
%

\vspace{-3mm}
\section{Experiment}
\vspace{-3mm}
The purpose of the experiment is to assess the effectiveness of the Two-stage Predict and then Optimize approach in addressing the IRP. The IRP is a problem that can be formulated as a Mixed Integer Linear Programming (MILP) model. For our experiment, we will focus on the standard MILP formulation.
\begin{equation}
\vspace{-1mm}
 \min c^{\top} x,  \text{subject to } A x = b,  x \geq 0 \quad \text{some or all } x_i \text{ integer}, \text{where } c \in \mathbb{R}^k, A \in \mathbb{R}^{p \times k}, b \in \mathbb{R}^p
\end{equation}
The optimal solution is defined  as $x^*(c ; A, b)$
\vspace{-3mm}
\subsection{Regret In Two-Stage Approach}
\vspace{-3mm}
In two-stage Predict then Optimize settings, the customer's demand of coefficient $(c)$ parameter is unknown. We have incorporated customer demand-related features and leveraged a neural network model to predict customer demand. The aim of employing this model is to calculate the regret, which is defined as the difference between the objective value using the predicted value from the model and the actual value, for different prediction accuracy, i.e., $\operatorname{regret}(\hat{c}, c ; A, b)=c^{\top}\left(x^*(\hat{c} ; A, b)-x^*(c ; A, b)\right)$.
\vspace{-7mm}
\begin{figure}[H]
\centering
\includegraphics[scale=0.4]{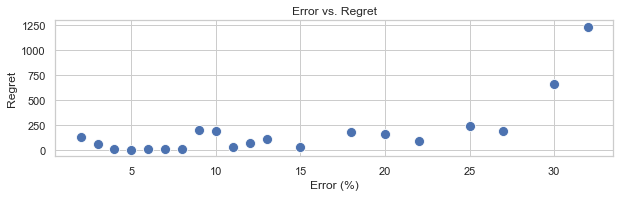}
\caption{Regret vs ML Model Error}
\vspace{-3mm}
\label{fig:1-formulation}
\end{figure}
\vspace{-6mm}
The concept of regret in relation to prediction accuracy in Figure \ref{fig:1-formulation} indicates that solely training to minimize machine learning (ML) loss is not an effective strategy for optimizing decision-making in the context of the Inventory Routing Problem (IRP).
\vspace{-4mm}

\section{Model}
\begin{figure}[pht] 
\vspace{-5mm}
\centering {\includegraphics[scale=0.4]{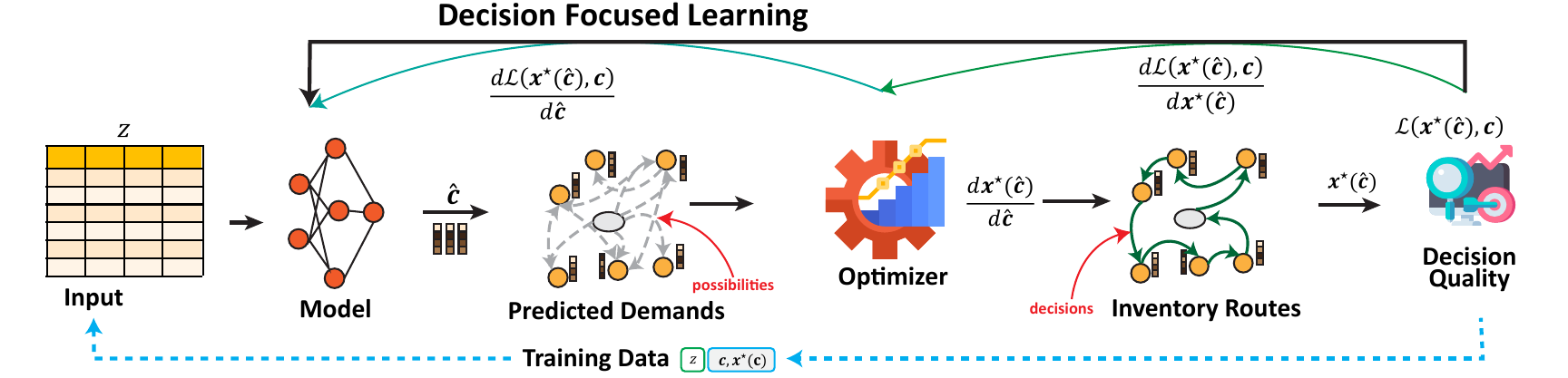}}
\caption{\textbf{Decision Focused Learning Approach Algorithm}.}\label{fig:2-model}
\vspace{-3mm}
\end{figure}

\vspace{-3mm}
The result of our experiment led us to explore Decision focused learning approach in Inventory Routing Problem (IRP). We need to differentiate the task loss or regret \(\mathcal{L}\) with respect to the model parameter \(\theta\). This can be calculated by; 
\begin{equation}
\vspace{-1mm}
\frac{\partial \mathcal{L}}{\partial \theta} = \frac{\partial \mathcal{L}}{\partial x^*(\hat{c} ; A, b)} \frac{\partial x^*(\hat{c} ; A, b)}{\partial \hat{c}} \frac{\partial \hat{c}}{\partial \theta}
\end{equation}
The first term is the gradient of the task loss with respect to the variable, and for IRP, $\frac{\partial L}{\partial x^*(\hat{c}; A, b)}=c$, and the third term is the gradient of the predictions with respect to the model parameter, $\frac{\partial \hat{c}}{\partial \theta}$, easily calculated using a Machine Learning model. The problem is finding,$\frac{\partial x^*(\hat{c} ; A, b)}{\partial \hat{c}}$, the gradient of the optimal solution $x^*(\hat{c} ; A, b)$ with respect to the parameter $\hat{c}$. Figure \ref{fig:2-model} depicts the overall approach of of decision focused learning in the context of IRPs.

\vspace{-3mm}
\subsection{Differentiable MILP}
\vspace{-3mm}
In decision-focused learning, making Linear programs twice differentiable is a common challenge. The objective function of IRP is linear, it is not twice differentiable with respect to prediction, to solve this issue the standard practice is to convert MILP into a Quadratic program(QP), and the gradient of Qp is computable through implicit differentiation of the KKT conditions. We can use the regularized term which transforms the linear program into a standard quadratic program (QP) as mentioned in \cite{wilder2019melding}. The Hessian of the QP is smaller than zero which ensures that QP provides a differentiable surrogate and we can use the standard QP solver \cite{amos2017optnet} to solve this problem.\\
Minimize the total cost of holding inventory, routing, and regularization:
\begin{equation}
\vspace{-1mm}
\text{Minimize:} \quad \sum_{i \in I} \sum_{t \in T} (h_s \cdot S_t + h_i \cdot s_{it}) + \sum_{t \in T} \sum_{k \in K} c_{kj} \cdot z_{kt} + \lambda \sum_{t \in T} \sum_{k \in K} x_{kt}^2
\vspace{-1mm}
\end{equation}
\vspace{-5mm}
\subsection{Interior Point Solving}
\vspace{-3mm}
Instead of adding the regularizer adding a logarithmic barrier can be another option. The log-barrier term is the most natural tool in LP problems as the KKT conditions of the log-barrier function define the primal and the dual constraints. This approach encourages solutions to remain within the feasible set while allowing for smooth transitions from relaxed (continuous) to integer solutions. We can use IntOpt solver \cite{mandi2020interior} for this method.
\begin{equation}
\vspace{-1mm}
\text{Minimize} \sum_{i \in I} \sum_{t \in T} (h_s \cdot S_t + h_i \cdot s_{it}) + \sum_{t \in T} \sum_{k \in K} c_{kj} \cdot z_{kt} + \mu \sum_{i \in I, t \in T} \left[\log(M_i - s_{it}) + \log(q_{it})\right]
\end{equation}
Where:- \(\mu\) is a positive parameter controlling the strength of the barrier term.\\
Both approaches involve solving the Linear Program (LP) to find the optimal solution \((x, y)\) and then computing \(\frac{\partial \mathcal{L}}{\partial \hat{c}}\) around it.
\vspace{-4mm}
\section{Discussion}
\vspace{-4mm}
 In our paper, we introduce a novel idea to tackle the intricate Inventory Routing Problems (IRP) using decision-focused learning.Our experiments highlight a critical insight: simply increasing model accuracy doesn't guarantee improved profitability or optimal decisions. This discovery has motivated us to embrace a decision-focused perspective. However, challenges emerge when calculating gradients for optimal solutions with respect to model parameters. To address this, we've introduced the concept of regularizers and logarithmic barriers. Traditional regularization demonstrates promise in smoothing the objective function, while the logarithmic barrier method proves useful for managing constraints. It's essential to recognize that IRPs are NP-hard and become particularly computationally demanding at a large scale. Finding optimal decisions through objective function differentiation can be counterproductive if the methodology isn't chosen wisely for objective function transformation. In the realm of IRP, the logarithmic barrier method stands out as a robust choice for efficiently handling capacity constraints. However,  to smoothen the objective and ensure differentiability for gradient-based optimization, regularization may be the more suitable path. Further research is imperative to determine the most effective approach among these two or any hybridized method for integrating decisions and predictions into a single model, ensuring the optimal selection of inventory and routing for establishing a resilient supply chain strategy.


\vspace{-4mm}
\section{Conclusion}
\vspace{-4mm}
The Inventory Routing Problem, despite its computational complexity, holds paramount significance as it harmoniously melds inventory management with vehicle routing decisions. In today's competitive business landscape, precision in every decision is imperative. Decision-focused learning emerges as a valuable asset, seamlessly uniting predictive models with decision-making, offering a more agile and responsive approach compared to the conventional method of independently solving prediction and prescription models in complex decision-making scenarios.

\setlength{\bibsep}{-0.5pt plus -0.1ex}
\footnotesize
\bibliography{paper}

\begin{thebibliography}{1}

\bibitem{amos2017optnet}
Brandon Amos and J~Zico Kolter.
\newblock Optnet: Differentiable optimization as a layer in neural networks.
\newblock In {\em International Conference on Machine Learning}, pages 136--145. PMLR, 2017.

\bibitem{campbell1998inventory}
Ann Campbell, Lloyd Clarke, Anton Kleywegt, and Martin Savelsbergh.
\newblock The inventory routing problem.
\newblock In {\em Fleet management and logistics}, pages 95--113. Springer, 1998.

\bibitem{mandi2020interior}
Jayanta Mandi and Tias Guns.
\newblock Interior point solving for lp-based prediction+optimisation.
\newblock {\em Advances in Neural Information Processing Systems}, 33:7272--7282, 2020.

\bibitem{vanderschueren2022predict}
Toon Vanderschueren, Tim Verdonck, Bart Baesens, and Wouter Verbeke.
\newblock Predict-then-optimize or predict-and-optimize? an empirical evaluation of cost-sensitive learning strategies.
\newblock {\em Information Sciences}, 594:400--415, 2022.

\bibitem{wilder2019melding}
Bryan Wilder, Bistra Dilkina, and Milind Tambe.
\newblock Melding the data-decisions pipeline: Decision-focused learning for combinatorial optimization.
\newblock In {\em Proceedings of the AAAI Conference on Artificial Intelligence}, volume~33, pages 1658--1665, 2019.

\end{thebibliography}
\vspace{-2mm}

\end{document}